\pdfoutput=1

\documentclass[11pt]{article}

\usepackage[]{EMNLP2022}

\usepackage{times}
\usepackage{latexsym}
\usepackage{algorithm,algorithmicx}
\usepackage{algpseudocode}
\usepackage{microtype}
\usepackage{booktabs}
\usepackage{multirow}
\usepackage{multicol}
\usepackage{array}
\usepackage{amsmath}
\usepackage{arydshln}
\usepackage{bbm}
\usepackage{graphicx}
\usepackage{graphics}
\usepackage{siunitx}
\usepackage{amsfonts}
\usepackage{makecell}
\usepackage{algorithm}

\usepackage{microtype}

\usepackage{inconsolata}

\usepackage{booktabs}
\usepackage{hyperref}
\usepackage{url}
\usepackage{graphicx}
\usepackage{xcolor}
\usepackage{amsmath}
\usepackage{amssymb}
\usepackage{color, colortbl}
\usepackage{makecell}
\usepackage{microtype}
\usepackage{xspace}

\usepackage[T1]{fontenc}

\usepackage[utf8]{inputenc}

\usepackage{microtype}

\usepackage{inconsolata}

\usepackage{times}
\usepackage{latexsym}
\usepackage{listings}
\usepackage[T1]{fontenc}

\usepackage[utf8]{inputenc}

\usepackage{microtype}

\usepackage{inconsolata}

\usepackage{booktabs}
\usepackage{hyperref}
\usepackage{url}
\usepackage{graphicx}
\usepackage{xcolor}
\usepackage{amsmath}
\usepackage{amssymb}
\usepackage{color, colortbl}
\usepackage{makecell}
\usepackage{microtype}
\usepackage{xspace}

\usepackage{array, tabularx, caption, boldline}
\usepackage{cellspace}
\usepackage{algpseudocode}  
\usepackage{amsmath}
\usepackage{multicol}  
\usepackage{multirow} 
\usepackage{flexisym}
\usepackage{graphicx}  
\usepackage{array}
\usepackage{multicol}
\usepackage{multirow}
\usepackage{amsmath}
\usepackage{booktabs}
\usepackage{array}
\usepackage{soul}
\usepackage{vcell}
\usepackage{pbox}
\usepackage{enumitem}
\usepackage{graphics}
\usepackage{wrapfig}
\usepackage{makecell}

\usepackage{tabularx}
\makeatletter
\def\hlinewd#1{%
\noalign{\ifnum0=`}\fi\hrule \@height #1 %
\futurelet\reserved@a\@xhline}
\makeatother

\definecolor{darkgreen}{rgb}{0, 0.25, 0.}
\definecolor{darkblue}{rgb}{0, 0., 0.35}
\definecolor{LightCyan}{rgb}{0.88,1,1}

\title{Explanations from Large Language Models Make Small Reasoners Better}

\author{\textbf{Shiyang Li}\textsuperscript{1}, \textbf{Jianshu Chen}\textsuperscript{2}, \textbf{Yelong Shen}\textsuperscript{3}, \textbf{Zhiyu Chen}\textsuperscript{1}, \textbf{Xinlu Zhang}\textsuperscript{1} , \textbf{Zekun Li}\textsuperscript{1} \\ \textbf{Hong Wang}\textsuperscript{1}, \textbf{Jing Qian}\textsuperscript{1}, \textbf{Baolin Peng}\textsuperscript{3}, \textbf{Yi Mao}\textsuperscript{3}, \textbf{Wenhu Chen}\textsuperscript{4} and \textbf{Xifeng Yan}\textsuperscript{1} \\
  \textsuperscript{1}University of California, Santa Barbara \\
  \textsuperscript{2}Tencent AI Lab, \textsuperscript{3}Microsoft \\
    \textsuperscript{4}University of Waterloo, Vector Institute \\
  {\tt \{shiyangli,zhiyuchen,xinluzhang,zekunli,hongwang600,jing\_qian,xyan\}@cs.ucsb.edu} \\ 
   {\tt jianshuchen@tencent.com}, {\tt wenhuchen@uwaterloo.ca} \\
  {\tt \{yelong.shen,bapeng,maoyi\}@microsoft.com}}
\begin{document}
\maketitle
\begin{abstract}
Integrating free-text explanations to in-context learning of large language models (LLM) is shown to elicit strong reasoning capabilities along with reasonable explanations.
In this paper, we consider the problem of leveraging the explanations generated by LLM to improve the training of small reasoners, which are more favorable in real-production deployment due to their low cost. We systematically explore three explanation generation approaches from LLM and utilize a multi-task learning framework to facilitate small models to acquire strong reasoning power together with explanation generation capabilities. Experiments on multiple reasoning tasks show that our method can consistently and significantly outperform finetuning baselines across different settings, and even perform better than finetuning/prompting a 60x larger GPT-3 (175B) model by up to 9.5\% in accuracy. As a side benefit, human evaluation further shows that our method can generate high-quality explanations to justify its predictions, moving towards the goal of explainable AI. 
\end{abstract}

\section{Introduction}
Large language models (LLM) have achieved impressive results with in-context learning; by adding a few demonstrations as the prompts, they can solve unseen tasks without any parameter update \citep{Brown2020LanguageMA,Thoppilan2022LaMDALM,Chowdhery2022PaLMSL,wei2022finetuned}.
Recently, it is shown that adding explanation-augmented prompts can elicit strong performance in various reasoning tasks \citep{Wei2022ChainOT,Lampinen2022CanLM}, such as math word problem \citep{Cobbe2021TrainingVT}, symbolic reasoning \citep{Wei2022ChainOT}, numerical reasoning \citep{Zhou2022LeasttoMostPE}  and commonsense reasoning tasks \citep{talmor-etal-2019-commonsenseqa}. 
In addition, they also enable LLM to generate reasonable explanations to justify the reasoning outcomes.

In this paper, we consider the problem of leveraging these elicited explanations by LLM to improve the training of small reasoners. Small language models (SLM)\footnote{
We argue that small and large models are relative concepts. For the same model, it can be small or large depending on the context.
} could be more favorable over LLM in many real situations due to their low cost in both storage and computation. Nevertheless, one important open question is how to close the performance gap with respect to LLM on complicated reasoning tasks, as is observed in \citet{Zelikman2022STaRBR}, especially in few-shot settings \citep{Li2019TeachingPM}. 
Surprisingly, \citet{HaseZXB20} shows that using human-annotated explanations does not improve the performance compared to standard finetuning on T5 \citep{Raffel2019ExploringTL}.
One possible reason is that many human-annotated explanations collected via crowdsourcing  \citep{Wiegreffe2021TeachMT} could be logically inconsistent and grammatically incorrect \citep{Narang2020WT5TT}, which restricts the amount of available high-quality explanations. On the other hand, using explanation-augmented prompts enables LLM to automatically generate descent explanations \citep{Wiegreffe2021ReframingHC}, making it a plausible alternative to generate arbitrary amount of explanations. Therefore, a key question is: \textit{Can the explanations generated by LLM improve the reasoning capability of SLM?}

\begin{figure*}[t!]
  \centering
  \includegraphics[scale=0.48]{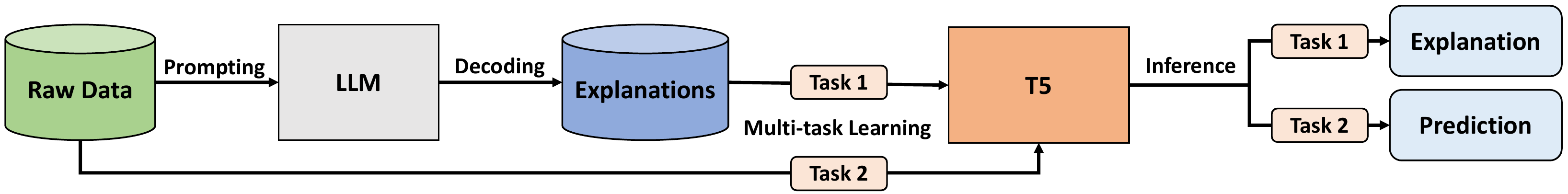}
  \caption{Overview of proposed framework.}
  \label{fig:pipeline}
\end{figure*}

In this paper, we show that explanations generated from LLM can consistently improve reasoning capability of SLM. Our framework is shown in Figure \ref{fig:pipeline}. Specifically, we first utilize several examples with human-written explanations as demonstrations of LLM and then generate explanations for \textit{training} set. We systematically explore three approaches to generating explanations. The first approach utilizes explanations generated through chain of thought prompting and explanations are adopted if LLM have correct predictions and are rejected otherwise \citep{Zelikman2022STaRBR}. The second one is to generate explanations by rationalization
prompting conditioned on golden labels \citep{Wiegreffe2021ReframingHC}. Intuitively, the first approach may generate higher quality explanations than the second if LLM' predictions are correct as incorrect explanations tend to generate incorrect predictions \citep{Wei2022ChainOT}. However, the first approach will reject explanations on problems with incorrect predictions, leaving their explanations empty. On the other hand, the second one explicitly conditions on golden labels and may still generate useful explanations on problems where chain of thought prompting cannot predict correctly. Therefore, we propose a third hybrid approach: adopting explanations generated by chain of thought prompting if LLM have correct predictions and use rationalization prompting otherwise. As we will show in section \ref{experiments}, three explanation generation methods can consistently and significantly improve fine-tuning baselines without explanations and our hybrid approach achieves best results on two of three datasets.

\begin{figure*}[t!]
  \centering
  \includegraphics[scale=0.52]{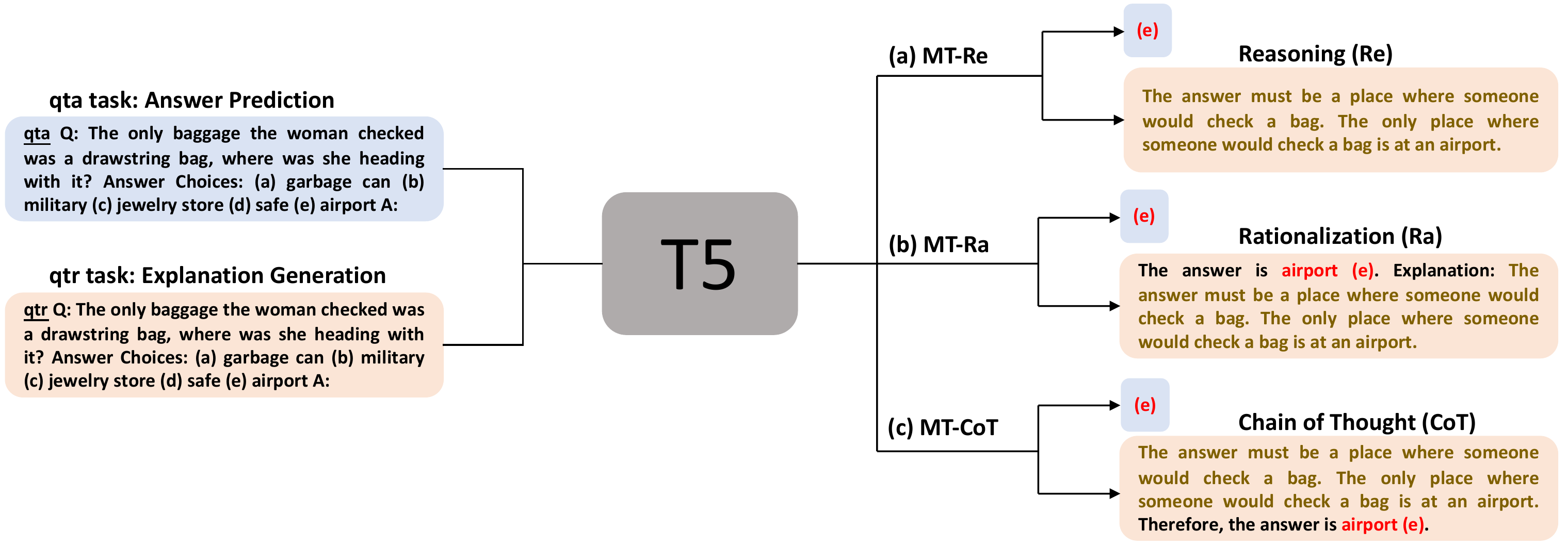}
  \caption{The comparison among (a) MT-Re \citep{HaseZXB20}, (b) MT-Ra \citep{Camburu2018eSNLINL} and (c) our proposed MT-CoT for multi-task learning with explanations under text-to-text format using T5. Left parts are inputs of T5 and right parts are targets for different multi-task learning setups. Task \textit{qta} (question to answer) is trained to directly generate answers for all three modes while \textit{qtr} (question to reason) task is trained to generate reasoning, rationalization and chain of thought for (a) MT-Re, (b) MT-Ra and (c) MT-CoT, respectively.}
  \label{fig:mt_example}
\end{figure*}

We further adopt a multi-task (MT) learning framework shown in Figure \ref{fig:mt_example} to utilize the LLM-generated explanations since (1) it can naturally allow training with partially generated explanations and (2) self-rationalizing model \citep{Wiegreffe2021MeasuringAB}, where golden label and the human-written explanation is linearly concatenated as the target, performs significantly worse than MT counterpart\citep{HaseZXB20}. Interestingly, we find that even with the same MT approaches (i.e., MT-Re \citep{HaseZXB20} and MT-Ra \citep{Camburu2018eSNLINL} ) as \citet{HaseZXB20}, we can consistently and significantly improve strong T5 standard finetuning baseline using LLM-generated explanations, which is in stark contrast to the results in \citet{HaseZXB20}, where finetuning T5 with MT-Re and MT-Ra only achieves on par results using crowdsourced ones. In addition, we further propose MT-CoT, where the small language model is trained to jointly solve two tasks: (i) directly generating the answer and (ii) generating an explanation and then the answer, as shown in Figure \ref{fig:mt_example} (c). Unlike MT-Re and MT-Ra, MT-CoT positions the answer after the explanation, hoping the model can learn to derive it from the explanation like chain of thought \cite{Wei2022ChainOT}. Our results show that all three explanation generation approaches can improve reasoning capability of small language models with MT-Ra, MT-Re and MT-CoT setups. And MT-CoT can achieve the best results over MT-Re and MT-Ra on two of three datasets. In addition, our method can outperform standard finetuning baseline by up to 8.1\% in accuracy and even perform better than finetuning/prompting a 60x larger GPT-3 model (175B) by up to 9.5\% in accuracy on \textit{CommonsenseQA}. Finally, as a side benefit, human evaluation further shows that our method can generate high-quality explanations to justify its predictions, moving towards the goal of explainable AI \citep{samek2019explainable}. 

In a nutshell, we summarize our contribution as following:
\begin{itemize}
\item We show that multi-task learning with explanations from LLM can consistently and significantly improve strong T5 single-task fine-tuning baselines across various settings. 
\item We propose a hybrid prompting approach to generating explanations from LLM and MT-CoT to further improve our learning with explanations from LLM paradiam.
\item We demonstrate that our method can perform better than finetuning/prompting a 60x larger GPT-3 model
(175B) by up to 9.5\% in accuracy on \textit{CommonsenseQA} and generate high-quality explanations to justify its predictions towards the goal of explainable AI. 
\end{itemize}
\section{Related Work}
 \label{sec:explanation_generation}
 
\paragraph{Prompting with Explanations.} Recently, a new learning paradigm, \textit{in-context learning} where several training examples are used as demonstrations of LLM without any parameter update, has shown promising results in various NLP tasks \citep{Brown2020LanguageMA}. Although promising, LLM still struggle with tasks requiring strong reasoning capability \citep{Wei2022ChainOT}. To enable better few-shot in-context learning of LLM for reasoning tasks, \citet{Wei2022ChainOT} proposes chain of thought prompting, which provides intermediate reasoning steps as explanations in prompts before answers and has achieved state-of-the-art in arithmetic, symbolic and common sense reasoning tasks. \citet{Zhou2022LeasttoMostPE} further extends chain of thought prompting with least-to-most
prompting, which decomposes a complex problem into a list of subproblems with natural languages, and then sequentially solves these subproblems in a recursive fashion.  \citet{Kojima2022LargeLM} moves one step further and shows that LLM are zero-shot reasoners by simply adding “\textit{Let’s think step by step}” without any demonstration in prompts. Unlike these work, \citet{Lampinen2022CanLM} explores explanations after answers prompting for LLM, where answers are fed into LLM before providing their explanations in prompts, and also observes consistent gains. 

These also exist work to utilize explanations generated from LLM rather than focusing on their final predictions. \citet{Wiegreffe2021ReframingHC} explores utilizing LLM to annotate explanations for existing datasets and proposes a sample-then-filter paradigm with human annotations. \citet{Ye2022TheUO} proposes to utilize a calibrator to calibrate GPT-3 as they find that GPT-3 tends to generate consistent but less factual explanations for textual
reasoning tasks. However, none of these work explores if these noisy explanations generated from LLM without human-involved filtering can be used to improve SLM reasoning capability. The closest work to ours is STaR \citep{Zelikman2022STaRBR}. STaR begins with prompting a descent large language model GPT-J with 6B parameters \citep{mesh-transformer-jax} possibly including answer hints via chain of thought prompting to generate explanations with incorrect answer rejection. After that, they utilize filtered training datasets with explanations to finetune GPT-J as a teacher model and then utilize the teacher model to generate explanations of training datasets to train a student GPT-J model iteratively with a self-training fashion until performance plateaus. However, STaR often requires dozens of iterations to converge, which is both time-consuming and compute-intensive to train a large 6B model. What's worse, their method may not be applicable to smaller language models, e.g. GPT-2 \citep{radford2019gpt2} and strong non-autoregressive models, e.g. T5, as they may not generate high-quality explanations with prompting. In addition, they only focus on chain of thought style prompting and finetuning while our approach can improve SLM across model sizes, explanation generation and multi-task finetuning methods.

\paragraph{Learning with Explanations.} Learning with explanations has been commonly studied in robotics \citep{johnson1994agents} and computer vision \citep{Hendricks2016GeneratingVE}. Recently, it has received increasing attention in NLP as well. \citet{Camburu2018eSNLINL} proposes MT-Ra for natural language inference task with LSTM and does not observe gains over single-task finetuning. \citet{Narang2020WT5TT} utilizes MT-Ra setup on both T5-base and T5-11B models but mainly focuses on explanation generation. Instead, \citet{rajani-etal-2019-explain} observes improvements with two-stage finetuning using human-annotated explanations for common sense reasoning task, where the first stage is to train a model for explanation generations with GPT \citep{Radford2018GPT} and the second one utilizes explanations as input to train a classification model based on BERT \citep{devlin2019bert}. However, \citet{HaseZXB20} finds that both two-stage finetuning and multi-task learning with MT-Re and MT-Ra setups only obtain comparable results over standard finetuning baselines on T5. We instead show that MT-Re, MT-Ra and our proposed MT-CoT with explanations from LLM can consistently and significantly outperform standard finetuning baselines without accuracy-explanation trade-off \citep{jain-etal-2020-learning}.

\section{Explanation Generation from LLM}
\noindent{\textbf{Problem setup.}} Denote $D =\{(x_{i}$, $y_{i})\}^{N}$ to be a dataset with $N$ training instances, where  $x_{i}$ is a problem and $y_{i}$ is its answer. Also, we have a handful of human-written instances $E =\{(x^{p}_{i}$, $e^{p}_{i}$ , $y^{p}_{i})\}^{M}$, where $e^{p}_{i}$ is a free-text explanation to explain why a problem $x^{p}_{i}$ has $y^{p}_{i}$ as its answer and $\{(x^{p}_{i}$, $y^{p}_{i})\}^{M} \subset D$ with $M \ll N$ (we set $M = 7$ in our experiments). Our goal is to fully leverage LLM with $E$ as demonstrations for in-context learning to generate explanation $e_{i}$ for all $(x_{i}$, $y_{i})$, where $1\leq i \leq N$, so that we can utilize these generated explanations from LLM to improve SLM reasoning capability.

\paragraph{COTE.} A chain of thought is a series of intermediate reasoning steps before providing an answer of a problem, mimicking human deliberate thinking process to perform complicated reasoning tasks \citep{Wei2022ChainOT}. Chain of thought prompting provides intermediate reasoning steps as explanations before answers in prompts. Formally, for $1\leq i \leq N$, we first concatenate all instances in $E$ and $x_{i}$ as prompt $\hat{p}_{i}$ = ($x^{p}_{1}$, $e^{p}_{1}$ , $y^{p}_{1}$, ..., $x^{p}_{M}$, $e^{p}_{M}$, $y^{p}_{M}$,  $x_{i}$). We then feed prompt $\hat{p}_{i}$ into LLM and greedily decode until a stop token is generated. After that, we parse the decoded sentence as explanation part $\hat{e}_{i}$ and prediction part $\hat{y}_{i}$. Intuitively, if $\hat{y}_{i} \neq y_{i}$, $\hat{e}_{i}$ may not have high quality as incorrect explanations tend to generate incorrect predictions \citep{Wei2022ChainOT}. Thus, we utilize \textit{\textbf{C}hain \textbf{O}f \textbf{T}hought prompting with incorrect answer r\textbf{E}jection} (COTE) \citep{Zelikman2022STaRBR} by only adopting $e_{i} := \hat{e}_{i} $ if $\hat{y}_{i} = y_{i} $; otherwise, we reject $\hat{e}_{i}$ and set $e_{i}$ as \textit{none}.

\paragraph{RP.} Since COTE uses the answers in original datasets to reject explanations with incorrect predictions, these instances will no longer have explanations. To alleviate this issue, an alternative is apply \textit{\textbf{R}ationalization \textbf{P}rompting }(RP) \citep{Wiegreffe2021ReframingHC} to generate explanations for every instance in training sets. Unlike COTE, RP provides explanations given golden answers. Specifically, for $1\leq i \leq N$, we concatenate all instances in $E$ and $(x_{i}, y_{i})$ as prompt $\bar{p}_{i}$ = ($x^{p}_{1}$, $y^{p}_{1}$, $e^{p}_{1}$ ,  ..., $x^{p}_{M}$,  $y^{p}_{M}$, $e^{p}_{M}$,  $x_{i}$, $y_{i}$). We then feed prompt $\bar{p}_{i}$ into LLM and greedily decode until a stop token is generated. The decoded sentence $\bar{e}_{i}$ is cast as explanation $\hat{e}_{i}$, i.e. $e_{i} := \bar{e}_{i}$, without filtering.

\paragraph{CROP.} COTE will possibly generate relatively high-quality explanations if LLM give correct predictions of problems at hand as incorrect explanations tend to generate incorrect predictions \citep{Wei2022ChainOT}. However, for problems with incorrect predictions, COTE casts their explanations as \textit{none}. On the other hand, RP can generate explanations for every instance in the dataset, but we cannot easily assess their quality without human annotation. Therefore, we propose \textit{\textbf{C}hain of Thought with \textbf{R}ationalization Pr\textbf{O}mpting backu\textbf{P}} (CROP), where when COTE generates \textit{none} as explanations, we will utilize RP as a backup approach. Intuitively, if LLM cannot predict a problem correctly under chain of thought prompting, the problem may be difficult \citep{Zelikman2022STaRBR} and RP may provide a meaningful explanation as it can access golden label during explanation generation process.

\section{Multi-task Learning with Explanations}
\label{sec:mt_explanation}

In this section, we elaborate how to utilize explanations generated from LLM to improve SLM reasoning capability with a multi-task learning framework. We detail three multi-task learning with explanations methods in the following.

\paragraph{MT-Re.} Multi-task Learning with Reasoning (MT-Re) is introduced by \citet{HaseZXB20} (see Figure \ref{fig:mt_example} (a)). MT-Re is trained to directly generate predictions for \textit{qta} (question to answer) task the same as standard finetuning without explanations and generate explanations without explicitly providing answers in \textit{qtr} (question to reason) task. The training objective of MT-Re is to mix loss $\mathcal{L_{\text{qta}}}$ for \textit{qta} task and $\mathcal{L_{\text{qtr}}}$ for \textit{qtr} task: 
\begin{equation}\label{eq:mt_obj}
  \mathcal{L_{\text{mt}}}  = \alpha \mathcal{L_{\text{qta}}} + (1-\alpha) \mathcal{L_{\text{qtr}}} ,
\end{equation}
where $\alpha$ weights $\mathcal{L_{\text{qta}}}$ and $\mathcal{L_{\text{qtr}}}$ loss, and is tuned on development set.

\paragraph{MT-Ra.} Multi-task Learning with Rationalization (MT-Ra) is first proposed by \citet{Camburu2018eSNLINL} for natural language inference task using LSTM-based models \citep{hochreiter1997long} and we adopt it with a more powerful T5 model for other reasoning tasks. As shown in Figure \ref{fig:mt_example} (b), models are trained to generate predictions for \textit{qta} task the same as MT-Re and also trained to generate rationalization for \textit{qtr} task. This is different from MT-Re as MT-Ra allows explanations to be explicitly conditioned on predictions. For MT-Ra, we use the same training objective as Equation \ref{eq:mt_obj} and tune $\alpha$ on development set.

\paragraph{MT-CoT.} MT-Re does not explicitly model interactions between explanations and answers during training, which may make models hard to capture their relations. While MT-Ra is explicitly trained to generate explanations conditioned on answers, it may still have difficulty in understanding their causal effects as answers are never trained to explicitly access their explanations. To bridge this gap, we propose Multi-task Learning with Chain of Thought (MT-CoT), where models are trained to generate answers for \textit{qta} task and generate chain of thought for \textit{qtr} task, as shown in Figure \ref{fig:mt_example} (c). For MT-CoT, we use the same training objective as Equation \ref{eq:mt_obj} and tune $\alpha$ on development set.

In MT-CoT training paradigm, models not only know answers from \textit{qta} task but also are explicitly shown how answers are derived with intermediate reasoning steps before knowing them from \textit{qtr} task. As we will show in experiments, this training paradigm is a supplement to MT-Re and MT-Ra, and can consistently improve small language model reasoning capability and also outperform MT-Re and MT-Ra on two datasets. 

\section{Experiments}\label{experiments}

\begin{table*}[!h]
    \normalsize
	\centering 
	\renewcommand{\arraystretch}{1.2}
	\setlength{\tabcolsep}{4pt}
	\scalebox{0.935}{
	\begin{tabular}{ccccccccccc}
		\hlinewd{0.75pt}
 &\multicolumn{3}{c}{\textbf{CommonsenseQA}}&\multicolumn{3}{c}{\textbf{StrategyQA}}&\multicolumn{3}{c}{\textbf{OpenbookQA}}\\
		\cmidrule(lr){2-4}
		\cmidrule(lr){5-7}
		\cmidrule(lr){8-10}
		&COTE&RP&CROP&COTE&RP&CROP&COTE&RP&CROP\\
		\hline
        ST &\multicolumn{3}{c}{$63.05_{0.50}$}&\multicolumn{3}{c}{$58.60_{1.36}$}&\multicolumn{3}{c}{$58.08_{0.65}$}\\

		 MT-Re&$63.78_{0.43}$&$63.78_{0.20}$&$64.05_{0.22}$&$60.26_{0.92}$&$60.52_{0.81}$&$60.26_{0.62}$&$59.48_{0.93}$&$60.44_{1.49}$&$59.04_{1.63}$\\
		 
		 MT-Ra&$\underline{64.05}_{0.60}$&$\underline{64.14}_{0.22}$&$\underline{\textbf{64.50}}_{0.22}$&$\underline{60.52}_{0.86}$&$\underline{60.79}_{0.43}$&$60.61_{0.64}$&$58.68_{2.11}$&$59.52_{0.20}$&$\underline{60.40}_{0.59}$\\
		 
		 MT-CoT & $63.88_{0.14}$ &$63.69_{0.30}$&$63.75_{0.51}$&$60.26_{1.46}$&$\underline{60.79}_{1.31}$&$\underline{\textbf{61.05}}_{0.85}$&$\underline{\textbf{60.68}}_{0.37}$&$\underline{60.64}_{0.66}$&$59.64_{0.90}$\\		 

		\hlinewd{0.75pt}
	\end{tabular}}
    \caption{Accuracy comparison (\%) of single-task finetuning baselines (ST) with MT-Re, MT-Ra and MT-CoT utilizing explanations generated by COTE, RP and CROP. Results are averaged over five runs with their standard deviation in the subscript. Best results for each \underline{column} with the same explanations are underlined and best results for each \textbf{dataset} are bold.}
	\label{tab:main_results}
\end{table*}
\subsection{Experimental setup}
\label{sec:exp_setup}
We evaluate our methods on three reasoning tasks.

(1) \textbf{CommonsenseQA} \citep{talmor-etal-2019-commonsenseqa} is a 5-way multi-choice question answering dataset that requires common sense reasoning with 9741/1221/1140 for training/development/test set questions, respectively. Since its test set is not publicly available, we report results on its development set following previous work \citep{Zelikman2022STaRBR,Li2019TeachingPM}.

(2) \textbf{StrategyQA} is a binary yes/no question answering dataset requiring implicit multi-hop reasoning steps and
should be inferred using a strategy \citep{geva-etal-2021-aristotle}. It has 2290 training set and 490 test set questions. Since its test set is not publicly available, we utilize their split in GitHub \footnote{\url{https://github.com/eladsegal/strategyqa}}, where original training set is randomly split into 90\% for training and 10\% for development set. In our experiments, we report results on their Github development set and utilize their Github training set for training without utilizing explanations from their original annotations. 

(3) \textbf{OpenbookQA} is a 4-way multi-choice question answering dataset requiring open book facts with broad common knowledge and multi-hop reasoning \citep{mihaylov-etal-2018-suit}. It has 4957/500/500 questions for training/development/test set split, respectively and we report results on its test set.

\paragraph{Explanation generation from LLM} We utilize GPT-3 \textit{text-davinci-002} engine with official OpenAI API \footnote{\url{https://beta.openai.com/docs/models/gpt-3}} to generate explanations through greedy decoding (by setting temperature as 0) following in-context learning paradigm. In each dataset, we have the same 7-shot examples with human-written explanations for COTE, RP and CROP detailed in section \ref{sec:explanation_generation}. We defer details of prompts into Appendix \ref{appendix:prompt_details}. 

\paragraph{Multi-task learning with explanations.} After obtaining explanations by COTE, RP and CROP, we utilize MT-Re, MT-Ra and MT-CoT introduced in section \ref{sec:mt_explanation} to train models with explanations based on T5. We implement multi-task learning framework with Huggingface \textit{transformers} library \citep{wolf-etal-2020-transformers}. For baselines, we utilize single-task finetuning (ST) without explanations. For fair comparison with ST, we keep hyper-parameters of multi-task learning the same as its corresponding ST except weight $\alpha$ which we tune with grid search $\{0.1, 0.2, 0.3, 0.4, 0.5, 0.6, 0.7, 0.8, 0.9\}$ on development sets. When training on \textit{none} explanations generated by COTE, we mask their loss for \textit{qtr} task. For both ST and multi-task finetuning, we directly generate predictions from \textit{qta} task for fair comparisons.
\subsection{Main results}

\label{sec:main_results}

In this section, we compare results between multi-task learning with explanations and its single-task finetuning counterpart using full training data on three datasets introduced in section \ref{sec:exp_setup}. Specifically, we generate explanations for each dataset with COTE, RP and CROP, and for each explanation generation method, we train T5-base model under MT-Re, MT-Ra and MT-CoT setups with 5 different runs in each setting. For single-task finetuning baseline, we only keep \textit{qta} task by removing \textit{qtr} task in multi-task learning setup. Results are summarized in Table \ref{tab:main_results}.

Three multi-task learning with three different explanation generation methods consistently and significantly outperform single-task finetuning baselines, showing the effectiveness of utilizing explanations from LLM. However, MT-CoT and MT-Ra have 4 and 6 underlined results, respectively, while MT-Re does not have any. We hypothesize it is because MT-CoT and MT-Ra \textit{explicitly} mention answers by \textit{the answer is} in \textit{qtr} task, making it easier for T5 to model relations between explanations and answers. Considering best results for each dataset, two of three are obtained via CROP with the remaining one obtained by COTE, showing that chain of thought prompting generates better explanations for SLM finetuning when their predictions are correct and RP backup can possibly further improve SLM reasoning capability. In addition, two of these three best results are obtained by MT-CoT, demonstrating that our method MT-CoT can serve as a good candidate to improve SLM reasoning with explanations from the toolbox.

\label{sec:few_shot}
\begin{table}[!h]
    \small
	\centering 
	\renewcommand{\arraystretch}{1.2}
	\setlength{\tabcolsep}{6pt}
	\scalebox{1.0}{
	\begin{tabular}{ccccc}
		\hlinewd{0.75pt}
		&50&100&200&400\\
		\hline
		\multicolumn{5}{c}{CommonsenseQA} \\
		 ST & $21.92_{1.57}$ &$27.06_{2.83}$&$28.04_{2.78}$&$44.49_{2.16}$\\
		 MT&$\textbf{29.25}_{3.03}$&$\textbf{33.28}_{3.53}$&$\textbf{36.13}_{5.29}$&$\textbf{46.55}_{1.53}$\\
		 $\alpha$* & 0.1 & 0.2 & 0.3 & 0.6 \\
		 \hline
		 \multicolumn{5}{c}{OpenbookQA} \\
		 ST & $27.08_{2.96}$ &$28.32_{2.88}$&$30.68_{2.10}$&$37.80_{4.64}$\\
		 MT&$\textbf{29.76}_{3.74}$&$\textbf{32.92}_{0.95}$&$\textbf{34.84}_{1.27}$&$\textbf{43.68}_{0.94}$\\
		 $\alpha$* & 0.1 & 0.1 & 0.2 & 0.2  \\
		\hlinewd{0.75pt}
	\end{tabular}}
    \caption{Accuracy comparison (\%) between single-task finetuning (ST) and multi-task learning with explanations (MT) along with optimal $\alpha$* in development sets under different training sample sizes. Results are averaged over five different training data splits with their standard deviation listed in the subscript.}
	\label{tab:few_shot_results}
\end{table}

\subsection{Few-shot learning results}

We have shown the effectiveness of our method on full-training settings in section \ref{sec:main_results} and further explore if explanations can improve SLM reasoning capability under few-shot settings. We conduct few-shot learning experiments for both \textit{CommonsenseQA} and \textit{OpenbookQA} datasets with best settings in section \ref{sec:main_results}. Specifically, we choose MT-Ra finetuning with explanations generated by CROP for \textit{CommonsenseQA} dataset and MT-CoT finetuning with explanations generated by COTE for \textit{OpenbookQA} dataset. We conduct experiments with $\{50, 100, 200, 400\}$ training sample sizes for both datasets on T5-base model and for each sample size, we randomly sample five data splits from its whole training set and each data split has a single run. Similar to previous experiments, we have single-task finetuning as our baselines and tune $\alpha$ using grid search on development sets for multi-task learning experiments. Besides accuracy, we also report optimal $\alpha$ on development sets, denoted as $\alpha$*. Intuitively, if $\alpha$* is small, $\mathcal{L_{\text{qtr}}}$ loss has more weight in the multi-task learning training objective 
listed in Equation \ref{eq:mt_obj} and hence, explanations are more important for correct prediction. We summarize our results in Table \ref{tab:few_shot_results}.
 
Multi-task learning with explanations (MT) consistently and significantly outperforms single-task finetuning baselines (ST). For \textit{CommonsenseQA} dataset, when training sample sizes are in $\{50, 100, 200\}$, MT significantly improves over ST about 6\%-8\% absolute accuracy. For \textit{OpenbookQA} dataset, when training sample sizes are in $\{100, 200, 400\}$, MT improves over ST about 4\%-6\% absolute accuracy. More interestingly, $\alpha$* tends to be smaller when less training data is used on both datasets. Intuitively, when training data sizes are small, models may have difficulty in learning just from limited problem and answer pairs and hence, requires a small $\alpha$* in the multi-task training objective \ref{eq:mt_obj}, i.e. larger weight on $\mathcal{L_{\text{qtr}}}$ loss during multi-task learning process. These consistent and significant gains show that our method not only can improve results in full-training settings but also is very useful when training data is limited.

\begin{table}[!h]
    \small
	\centering 
	\renewcommand{\arraystretch}{1.2}
	\setlength{\tabcolsep}{8pt}
	\scalebox{1.0}{
	\begin{tabular}{ccccc}
		\hlinewd{0.75pt}
		&T5-small &T5-base &T5-large &T5-3B \\
		\hline
		\multicolumn{5}{c}{CommonsenseQA} \\
		 ST & $48.26 $  &$ 63.05 $&$ 72.56 $&${81.82}$\\ 
		 MT&$\textbf{49.17} $&$\textbf{64.50}$&$\textbf{74.37}$& \textbf{82.47} \\
		 \hline
		 \multicolumn{5}{c}{OpenbookQA} \\
		 ST & $50.36 $ &$ 58.08 $&$ 61.60 $&$ 76.60 $\\
		 MT&$\textbf{51.72}$&$\textbf{60.68}$&$\textbf{64.60}$&$\textbf{78.60}$\\
		\hlinewd{0.75pt}
	\end{tabular}}
    \caption{Accuracy comparison (\%) between ST and MT across different model sizes.}
	\label{tab:model_size_results}
\end{table}

\subsection{Results across model sizes}

\label{sec:model_sizes}

Previous experiments utilize T5-base model and we further explore if explanations can improve language model reasoning capability across model sizes. We conduct full-training set experiments for both \textit{CommonsenseQA} and \textit{OpenbookQA} datasets with best settings for each dataset in section \ref{sec:main_results} across $\{\text{T5-small}, \text{T5-base}, \text{T5-large}, \text{T5-3B}\}$. For T5-small and T5-base, we have five different runs for each setting and their average results are reported. For T5-large and T5-3B, we only report a single run due to their intensive computational cost. Results are summarized in Table \ref{tab:model_size_results}.

MT consistently improves its ST counterpart on both \textit{CommonsenQA} and \textit{OpenbookQA} across model sizes from T5-small (60 million parameters) to T5-3B. For \textit{CommonsenQA}, MT improves ST about  0.7\%-1.8\% absolute  accuracy and for \textit{OpenbookQA}, MT improves ST about 1.4\%-3.0\% absolute  accuracy. Even for T5-3B, MT can improve strong ST with 2\% absolute  accuracy. These consistent results show that our approach can work on both small and relatively large models.   

\begin{table}[!h]
    \small
	\centering 
	\renewcommand{\arraystretch}{1.2}
	\setlength{\tabcolsep}{5pt}
	\scalebox{0.85}{
	\begin{tabular}{ccccc}
		\hlinewd{0.75pt}
		&CSQA & OBQA\\
		\hline
		 GPT-J Direct Finetuning (6B) $\diamond$ & 60.0 & - \\
		 STaR (6B) $\diamond$ & 72.5 & - \\
		 GPT-3 Direct Finetuning (175B)*  & 73.0 & - \\
		 GPT-3 Direct Prompting (175B)  & 80.59 & 83.00 \\
		 GPT-3 Chain of Thought Prompting (175B)   & 73.71 & 72.60 \\
		 GPT-3 Explain. after Answers Prompting (175B) & 80.84 & \textbf{83.40} \\
		 \hline
		 T5 MT (3B) & \textbf{82.47} & 78.60 \\
		\hlinewd{0.75pt}
	\end{tabular}}
    \caption{Accuracy comparison (\%) between T5 multi-task learning with explanations with various state-of-the-art LLM on CommonsenseQA (CSQA) and OpenbookQA (OBQA), and model sizes are listed in the parenthesis. Results with $\diamond$ and  * are from \citet{Zelikman2022STaRBR} and \citet{Xu2021HumanPO}, respectively.}
	\label{tab:llm_compare}
\end{table}
\subsection{Comparison with Large Language Models}
\begin{figure*}[!t]
  \centering
  \includegraphics[scale=0.48]{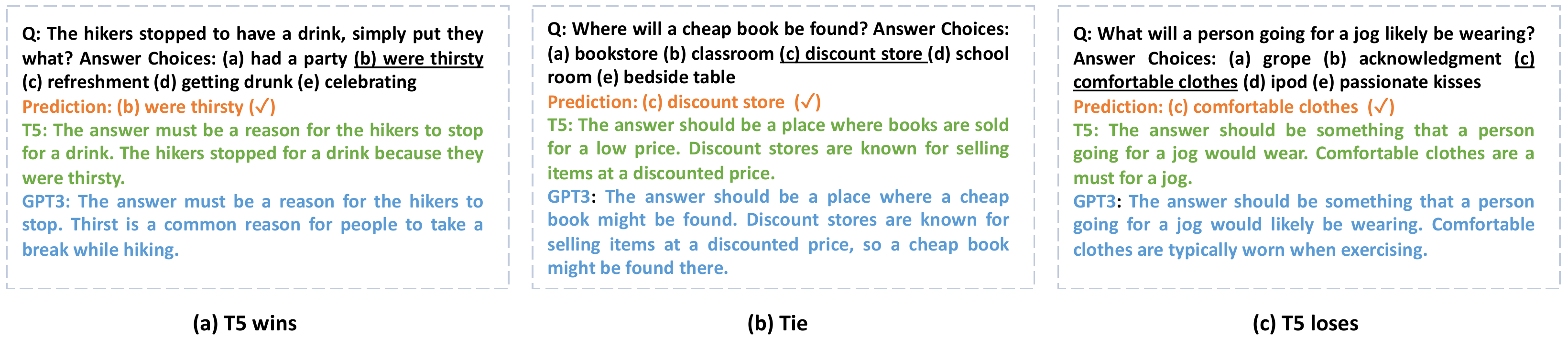}
  \caption{T5 and GPT-3 generated explanations used in human evaluation.}
    \label{fig:human_eval_example}
\end{figure*}
We further compare our method on T5-3B with state-of-the-art LLM. Specifically, we adopt GPT-J direct finetuning, its self-bootstrapping version (STaR) \citep{Zelikman2022STaRBR} and GPT-3 direct finetuning \citep{Xu2021HumanPO} as baseline methods with parameter update on downstream tasks. We also adopt GPT-3 direct prompting \citep{Brown2020LanguageMA},  GPT-3 chain of thought prompting \citep{Wei2022ChainOT} and GPT-3 explanations after answers prompting \citep{Lampinen2022CanLM} as prompting baselines. These three prompting methods utilize the same set of demonstrations for explanation generation in section \ref{sec:explanation_generation} and we defer their prompts into Appendix \ref{appendix:prompt_details}. Results are summarized in Table \ref{tab:llm_compare}.

Our approach can outperform strong 60x larger GPT-3 finetuning and various GPT-3 prompting methods on \textit{CommonsenseQA} up to about 9.5\% absolute accuracy. Also, although STaR can outperform its GPT-J baseline with chain-of-thought style iterative finetuning, their result still has about 10\% absolute accuracy gap with our method on \textit{CommonsenseQA} even with doubled parameter size and more compute during iterative finetuning process. For \textit{OpenbookQA}, our model underperforms GPT-3 direct prompting and explanations after answers prompting but can still outperform GPT-3 chain of thought prompting with 6\% absolute accuracy. In short, our method can achieve strong performance even compared with 60x larger GPT-3. 

\begin{table}[t!]
\centering
\small
\scalebox{0.95}{
\begin{tabular}{ccccccc}
\toprule
\multicolumn{3}{c}{Preference} & \vline & \multicolumn{3}{c}{Agreement Level} \\
\midrule
T5       & Tie      & GPT-3 & \vline   & Level 0  & Level 1 & Level 2 \\
14\%     & 44\%     & 42\%  & \vline   & 7\%      & 56\%     & 37\%   \\
\bottomrule
\end{tabular}}
    \caption{Head-to-head human explanation preference comparison between T5 (3B) and GPT-3 (175B) on \textit{CommonsenseQA} along with their agreement percentage on three levels.}
	\label{tab:human_eval}
\end{table}
\subsection{Human evaluation on generated explanations}
A side benefit of our model is to generate explanations towards more explainable AI to alleviate the notorious black box issue of deep neural networks \citep{Koh2017UnderstandingBP}. To evaluate quality of generated explanations from \textit{qtr} task for our model, we conduct human evaluation since automatic metrics are not highly correlated with human assessment \citep{clinciu-etal-2021-study,Kayser2021eViLAD}. 

Specifically, we perform a head-to-head explanation comparison on \textit{CommonsenseQA} dataset between T5-3B and GPT-3 175B few-shot explanations after answers
prompting since these models achieve close performance on this dataset, as shown in Table \ref{tab:llm_compare}. T5 model is trained with explanations generated by GPT-3 and we would like to know its generated explanation quality compared to that of GPT-3, which has been shown to be high-quality in \citet{Wiegreffe2021ReframingHC}. Therefore, we randomly sample 100 examples that are predicted correctly by both GPT-3 and T5, and for each example, we present a question, its ground truth answer and two randomly shuffle explanations as (a) and (b) generated by T5 and GPT-3 to three different human annotators with advanced NLP backgrounds and then ask them which explanation they prefer: (a), (b) or tie, similar to \citet{Wiegreffe2021ReframingHC}. Finally, we adopt a majority vote approach to decide preference on each example if at least two annotators have the same preference; otherwise, we cast that example's two explanations are tied. In addition, we report agreement percentage across three levels. Level 0 means all three annotators have different preferences, level 1 means only two annotators have the same preference and level 2 means all three annotators have the same preference. Results are summarized in Table \ref{tab:human_eval}.

As expected, explanations generated by T5 are less preferred over those from GPT-3 but there are still 58\% (14\%+44\%) explanations having better or competitive quality over GPT-3. In addition, more than 60\% explanations have disagreement (7\% in level 0 + 56\% in level 1). Given \citet{Wiegreffe2021ReframingHC} finds GPT-3 can generate competitive explanations even compared to human-written ones, we argue that this high disagreement is because explanations generated by both T5 and GPT-3 are high-quality, making humans hard to choose. To verify this hypothesis, we choose three T5 and GPT-3 generated explanation examples used in our human evaluation experiments,
as shown in Figure \ref{fig:human_eval_example}. Both T5 and GPT-3 can generate plausible explanations to justify their predictions and even though T5 loses to GPT-3 in example (c), its explanation is still reasonably good. We also provide examples with incorrect predictions in appendix \ref{appendix:examples}, some of which still have plausible predictions and explanations although different from golden labels. These results demonstrate explanations generated by our model have high quality even compared with strong GPT-3 with 60x larger size.

\section{Conclusion}

In this paper, we leverage explanations from LLM to improve small reasoners in a multi-task learning framework. Extensive experiments on multiple reasoning tasks show our method
can consistently and significantly outperform single-task finetuning baselines across explanation generation method, multi-task learning setups, training sample and small reasoner sizes, and can outperform strong finetuning/prompting a 60x larger GPT-3 on \textit{CommonsenseQA} by up to 9.5\% in accuracy. In addition, our model can generate high-quality explanations even compared to strong GPT-3 towards more explainable AI according to human evaluation.
\section*{Limitations}

Our approach requires a multi-task learning finetuning approach to integrate explanations into small language models and will require tuning weight $\alpha$ on a development set, which will require more compute during hyper-parameter tuning process. In addition, our work is constrained to textual reasoning problems, lacking more explorations in other reasoning tasks, e.g. symbolic reasoning and arithmetic reasoning, which we plan to leave as future work. 

\section*{Ethics Statement}
Our work is built on top of explanations generated from LLM, which have been observed to capture gender, race and religion biases \citep{Brown2020LanguageMA,lucy-bamman-2021-gender,Abid2021PersistentAB}. Generated explanations with these possible biases may be integrated into small models during finetuning process and be exposed when these small models generate explanations to justify their predictions. Therefore, our model could potentially share the same kinds of bias as the original LLM used for explanation generation. However, our multi-task learning framework naturally allows us to disable explanation generation and still enjoy performance gains by direct answer prediction without the risk of explicitly exposing these biases.

\section*{Acknowledgement}

This research was sponsored in part by the DARPA PTG program. Any opinions, findings, and conclusions or recommendations expressed in this material are those of the authors and do not necessarily reflect the views of the funding agencies.


\bibliography{anthology,custom}
\bibliographystyle{acl_natbib}


\newpage
\appendix
\section*{Appendix}
\section{Prompt details} \label{appendix:prompt_details}
Here we provides prompts we use in our experiments. Our prompts on \textit{CommonsenseQA} and \textit{StrategyQA} datasets are based on \citep{Zelikman2022STaRBR} and \citep{Wei2022ChainOT}, respectively. Explanations in prompts for \textit{OpenbookQA} are based on science facts in \textit{OpenbookQA} dataset Github repository \url{https://github.com/allenai/OpenBookQA}.

\begin{table}[h]
    \centering
    \small
    \begin{tabular}{p{7.8cm}}
        \toprule
        \vspace{-2mm}

Q: What do people use to absorb extra ink from a fountain pen?\\
Answer Choices:\\
(a) shirt pocket
(b) calligrapher's hand
(c) inkwell
(d) desk drawer
(e) blotter\\
\vspace{-1mm}
A: blotter (e)\\
~\\
Q: What home entertainment equipment requires cable?\\
Answer Choices:\\
(a) radio shack
(b) substation
(c) television
(d) cabinet
(e) desk\\
\vspace{-1mm}
A: television (c)\\
~\\
Q: The fox walked from the city into the forest, what was it looking for?\\
Answer Choices:\\
(a) pretty flowers
(b) hen house
(c) natural habitat
(d) storybook
(e) dense forest\\
\vspace{-1mm}
A: natural habitat (c)\\
~\\
Q: Sammy wanted to go to where the people were. Where might he go?\\
Answer Choices:\\
(a) populated areas
(b) race track
(c) desert
(d) apartment
(e) roadblock\\
\vspace{-1mm}
A: populated areas (a)\\
~\\
Q: Where do you put your grapes just before checking out?\\
Answer Choices:\\
(a) mouth
(b) grocery cart
(c) super market
(d) fruit basket
(e) fruit market\\
\vspace{-1mm}
A: grocery cart (b)\\
~\\
Q: Google Maps and other highway and street GPS services have replaced what?\\
Answer Choices:\\
(a) united states
(b) mexico
(c) countryside
(d) atlas
(e) oceans\\
\vspace{-1mm}
A: atlas (d)\\
~\\
Q: Before getting a divorce, what did the wife feel who was doing all the work?\\
Answer Choices:\\
(a) harder
(b) anguish
(c) bitterness
(d) tears
(e) sadness\\
\vspace{-1mm}
A: bitterness (c) \\
    \bottomrule
    \end{tabular}
    \caption{    
   Standard prompt for CommonsenseQA.
    }
\end{table}

\newpage

\begin{table}[H]

    \centering
    \small
    \begin{tabular}{p{7.8cm}}
        \toprule
        \vspace{-2mm}
        Q: What do people use to absorb extra ink from a fountain pen? \\
        Answer Choices:\\
        (a) shirt pocket (b) calligrapher's hand (c) inkwell (d) desk drawer (e) blotter \\
        \vspace{-1mm}
        A: The answer must be used to absorb extra ink. Blotters are designed to absorb liquids. Therefore, the answer is blotter (e).\\
         ~\\
Q: What home entertainment equipment requires cable? \\
Answer Choices: \\ (a) radio shack(b) substation (c) television (d) cabinet (e) desk\\
\vspace{-1mm}
A: The answer must require cable. Cable is used to provide satellite channels to televisions. Therefore, the answer is television (c). \\
~ \\  
Q: The fox walked from the city into the forest, what was it looking for? \\
Answer Choices:\\ (a) pretty flowers (b) hen house (c) natural habitat (d) storybook (e) dense forest \\
\vspace{-1mm}
A: The answer must be a reason for a fox to go into the forest. The forest is a fox's natural habitat. Therefore, the answer is natural habitat (c). \\
~\\
Q: Sammy wanted to go to where the people were. Where might he go? \\
Answer Choices:\\
(a) populated areas (b) race track (c) desert (d) apartment (e) roadblock\\
\vspace{-1mm}
A: The answer must be a place with many people. Populated areas, by definition, have a lot of people. Therefore, the answer is populated areas (a).\\
~\\
Q: Where do you put your grapes just before checking out?\\
Answer Choices:\\
(a) mouth (b) grocery cart (c) super market (d) fruit basket (e) fruit market\\
\vspace{-1mm}
A: The answer should be the place where grocery items are placed before checking out. Of the above choices, grocery cart makes the most sense for holding grocery items. Therefore, the answer is grocery cart (b).\\
~\\
Q: Google Maps and other highway and street GPS services have replaced what?\\
Answer Choices:\\ (a) united states (b) mexico (c) countryside (d) atlas (e) oceans\\
\vspace{-1mm}
A: The answer must be something that used to do what Google Maps and GPS services do, which is give directions. Atlases were also used to give directions. Therefore, the answer is atlas (d).\\
~\\
Q: Before getting a divorce, what did the wife feel who was doing all the work?\\
Answer Choices:\\ 
(a) harder (b) anguish (c) bitterness (d) tears (e) sadness\\
\vspace{-1mm}
A: The answer should be a feeling which would cause someone who was doing all the work to get divorced. If someone feels bitter towards their spouse, they are likely to want a divorce. Therefore, the answer is bitterness (c).\\
    \bottomrule
    \end{tabular}
        \caption{    
    Chain of Thought prompt for CommonsenseQA.
    }
\end{table}

\begin{table}[H]

    \small
    \centering
    \begin{tabular}{p{7.8cm}}
    \toprule
Q: What do people use to absorb extra ink from a fountain pen? \\
Answer Choices:\\
(a) shirt pocket
(b) calligrapher's hand
(c) inkwell
(d) desk drawer
(e) blotter \\
\vspace{-1mm}
A: blotter (e) \\
Explanation: the answer must be used to absorb extra ink. Blotters are designed to absorb liquids.\\
~\\
Q: What home entertainment equipment requires cable?\\
Answer Choices:\\
(a) radio shack
(b) substation
(c) television
(d) cabinet
(e) desk\\
\vspace{-1mm}
A: television (c) \\
Explanation: the answer must require cable. Cable is used to provide satellite channels to televisions. \\
~\\
Q: The fox walked from the city into the forest, what was it looking for?\\
Answer Choices:\\
(a) pretty flowers
(b) hen house
(c) natural habitat
(d) storybook
(e) dense forest\\
\vspace{-1mm}
A: natural habitat (c)\\
Explanation: the answer must be a reason for a fox to go into the forest. The forest is a fox's natural habitat.\\
~\\
Q: Sammy wanted to go to where the people were. Where might he go?\\
Answer Choices:\\
(a) populated areas
(b) race track
(c) desert
(d) apartment
(e) roadblock\\
\vspace{-1mm}
A: populated areas (a) \\Explanation: the answer must be a place with many people. Populated areas, by definition, have a lot of people.\\
~\\
Q: Where do you put your grapes just before checking out?\\
Answer Choices:\\
(a) mouth
(b) grocery cart
(c) super market
(d) fruit basket
(e) fruit market\\
\vspace{-1mm}
A: grocery cart (b) \\Explanation: the answer should be the place where grocery items are placed before checking out. Of the above choices, grocery cart makes the most sense for holding grocery items.\\
~\\
Q: Google Maps and other highway and street GPS services have replaced what?\\
Answer Choices:\\
(a) united states
(b) mexico
(c) countryside
(d) atlas
(e) oceans\\
\vspace{-1mm}
A: atlas (d) \\Explanation: the answer must be something that used to do what Google Maps and GPS services do, which is give directions. Atlases were also used to give directions.\\
~\\
Q: Before getting a divorce, what did the wife feel who was doing all the work? \\
Answer Choices:\\
(a) harder
(b) anguish
(c) bitterness
(d) tears
(e) sadness\\
\vspace{-1mm}
A: bitterness (c) \\Explanation: the answer should be a feeling which would cause someone who was doing all the work to get divorced. If someone feels bitter towards their spouse, they are likely to want a divorce.\\
\bottomrule
\end{tabular}
    \caption{
    Rationalization prompt for CommonsenseQA.
   }
\end{table}

\begin{table}[H]

    \centering
    \small
    \begin{tabular}{p{7.8cm}}
        \toprule
        \vspace{-2mm}
Q: Yes or no: Do hamsters provide food for any animals?\\
\vspace{-1mm}
A: Hamsters are prey animals. Prey are food for predators. Thus, hamsters provide food for some animals. So the answer is yes.\\
~\\
Q: Yes or no: Could the members of The Police perform lawful arrests? \\
\vspace{-1mm}
A: The members of The Police were musicians, not law enforcement officers. Only law enforcement officers can perform lawful arrests. Thus, the members of The Police could not perform lawful arrests. So the answer is no.\\
~\\
Q: Yes or no: Hydrogen's atomic number squared exceeds number of Spice Girls?\\
\vspace{-1mm}
A: Hydrogen has an atomic number of 1. 1 squared is 1. There are 5 Spice Girls. Thus, Hydrogen's atomic number squared is less than 5. So the answer is no.\\
~\\
Q: Yes or no: Is it common to see frost during some college commencements?\\
\vspace{-1mm}
A: College commencement ceremonies can happen in December, May, and June. December is in the winter, so there can be frost. Thus, there could be frost at some commencements. So the answer is yes. \\
~\\
Q: Yes or no: Could a llama birth twice during War in Vietnam (1945-46)?\\
\vspace{-1mm}
A: The War in Vietnam was 6 months. The gestation period for a llama is 11 months, which is more than 6 months. Thus, a llama could not give birth twice during the War in Vietnam. So the answer is no.\\
~\\
Q: Yes or no: Would a pear sink in water?\\
\vspace{-1mm}
A: The density of a pear is about 0.6 g/cm^3, which is less than water. Objects less dense than water float. Thus, a pear would float. So the answer is no.\\
~\\
Q: Yes or no: Is shrimp scampi definitely free of plastic?\\
\vspace{-1mm}
A: Shrimp scampi is a dish made with shrimp. Shrimp have been found to contain microplastics. Microplastics are plastic material. Thus, shrimp scampi may contain plastic. So the answer is no.\\
    \bottomrule
    \end{tabular}
    \caption{    
    Chain of Thought prompt for StrategyQA.
    }
\end{table}

\begin{table}[H]

    \centering
    \small
    \begin{tabular}{p{7.8cm}}
        \toprule
        \vspace{-2mm}
Q: Yes or no: Do hamsters provide food for any animals? \\
\vspace{-1mm}
A: yes\\
Explanation: Hamsters are prey animals. Prey are food for predators. Thus, hamsters provide food for some animals.\\
~\\
Q: Yes or no: Could the members of The Police perform lawful arrests?\\
\vspace{-1mm}
A: no\\
Explanation: The members of The Police were musicians, not law enforcement officers. Only law enforcement officers can perform lawful arrests. Thus, the members of The Police could not perform lawful arrests.\\
~\\
Q: Yes or no: Hydrogen's atomic number squared exceeds number of Spice Girls?\\
\vspace{-1mm}
A: no\\
Explanation: Hydrogen has an atomic number of 1. 1 squared is 1. There are 5 Spice Girls. Thus, Hydrogen's atomic number squared is less than 5.\\
~\\
Q: Yes or no: Is it common to see frost during some college commencements?\\
\vspace{-1mm}
A: yes\\
Explanation: College commencement ceremonies can happen in December, May, and June. December is in the winter, so there can be frost. Thus, there could be frost at some commencements.\\
~\\
Q: Yes or no: Could a llama birth twice during War in Vietnam (1945-46)?\\
\vspace{-1mm}
A: no\\
Explanation: The War in Vietnam was 6 months. The gestation period for a llama is 11 months, which is more than 6 months. Thus, a llama could not give birth twice during the War in Vietnam.\\
~\\
Q: Yes or no: Would a pear sink in water?\\
\vspace{-1mm}
A: no\\
Explanation: The density of a pear is about 0.6 g/cm^3, which is less than water. Objects less dense than water float. Thus, a pear would float.\\
~\\
Q: Yes or no: Is shrimp scampi definitely free of plastic?\\
\vspace{-1mm}
A: no\\
Explanation: Shrimp scampi is a dish made with shrimp. Shrimp have been found to contain microplastics. Microplastics are plastic material. Thus, shrimp scampi may contain plastic.\\
    \bottomrule
    \end{tabular}
    \caption{    
    Rationalization prompt for StrategyQA.
    }
\end{table}

\begin{table}[H]

    \centering
    \small
    \begin{tabular}{p{7.8cm}}
        \toprule
        \vspace{-2mm}
Q: What is the most likely to be an effect of acid rain on an aquatic environment?\\
Answer Choices:\\
(a) decrease in plant life
(b) increase in fish population
(c) increase in plant growth
(d) cleaner and clearer water\\
\vspace{-1mm}
A: (a) decrease in plant life\\
~\\
Q: The moon's surface\\
Answer Choices:\\
(a) is smooth on the entire surface
(b) contains large cavities cause by explosions
(c) contains an internal core of cheese
(d) is filled with lakes\\
\vspace{-1mm}
A: (b) contains large cavities cause by explosions\\
~\\
Q: As a car approaches you in the night\\
Answer Choices:\\
(a) the headlights become more intense
(b) the headlights recede into the dark
(c) the headlights remain at a constant
(d) the headlights turn off\\
\vspace{-1mm}
A: (a) the headlights become more intense\\
~\\
Q: When the weather changes as it does from Christmas to Easter,\\
Answer Choices:\\
(a) the air may chill
(b) the ground may freeze
(c) the plants may die
(d) the ground may warm\\
\vspace{-1mm}
A: (d) the ground may warm\\
~\\
Q: Heat and moisture in the ocean is a good recipe for\\
Answer Choices:\\
(a) a violent storm
(b) violent sea animals
(c) condensation
(d) inland storms\\
\vspace{-1mm}
A: (a) a violent storm\\
\vspace{-1mm}
Q: Hummingbirds take what with them\\
Answer Choices:\\
(a) Bees
(b) energy
(c) Pollen
(d) Honey\\
\vspace{-1mm}
A: (c) Pollen\\
~\\
Q: What covers over 90\% of the Earth's surface and 0\% of the moon's surface\\
Answer Choices:\\
(a) a magnesium iron silicate mineral
(b) chemical element with the symbol S
(c) the element with the symbol Fe
(d) that which contains 2 hydrogen and 1 oxygen molecules\\
\vspace{-1mm}
A: (d) that which contains 2 hydrogen and 1 oxygen molecules\\
    \bottomrule
    \end{tabular}
    \caption{    
   Standard prompt for OpenbookQA.
    }
\end{table}

\begin{table}[H]

    \centering
    \small
    \begin{tabular}{p{7.8cm}}
        \toprule
        \vspace{-2mm}
Q: What is the most likely to be an effect of acid rain on an aquatic environment? \\
Answer Choices:\\
(a) decrease in plant life (b) increase in fish population (c) increase in plant growth (d) cleaner and clearer water\\
\vspace{-1mm}
A: Acid rain has a negative impact on water quality and decrease in water quality leads to a decrease in aquatic life. So the answer is (a) decrease in plant life.\\
~\\
Q: The moon's surface\\
Answer Choices:\\
(a) is smooth on the entire surface (b) contains large cavities cause by explosions (c) contains an internal core of cheese (d) is filled with lakes\\
\vspace{-1mm}
A: The moon's surface contains many craters and craters are large cavities caused by explosions. So the answer is (b) contains large cavities cause by explosions.\\
~\\
Q: As a car approaches you in the night \\
Answer Choices: \\
(a) the headlights become more intense (b) the headlights recede into the dark (c) the headlights remain at a constant (d) the headlights turn off\\
\vspace{-1mm}
A: Headlights of a car are source of light. As a source of light becomes closer , that source will appear brighter. So the answer is (a) the headlights become more intense.\\
~\\
Q: When the weather changes as it does from Christmas to Easter,\\
Answer Choices:\\
(a) the air may chill (b) the ground may freeze (c) the plants may die (d) the ground may warm\\
\vspace{-1mm}
A: Christmas is in the winter season while Easter is in the beginning of spring. Weather change from winter to spring is to become warm. So the answer is (d) the ground may warm.\\
~\\
Q: Heat and moisture in the ocean is a good recipe for\\
Answer Choices:\\
(a) a violent storm (b) violent sea animals (c) condensation (d) inland storms\\
\vspace{-1mm}
A: An ocean is a source of heat and moisture for a hurricane and a hurricane is a violent storm. So the answer is (a) a violent storm.\\
~\\
Q: Hummingbirds take what with them\\
Answer Choices:\\
(a) Bees (b) energy (c) Pollen (d) Honey\\
\vspace{-1mm}
A: When hummingbirds dip their long bills into flowers to drink nectar, they also pick up grains of pollen. Pollen that sticks to a hummingbird’s feathers and bill gets carried to the next flower it visits. So the answer is (c) Pollen.\\
~\\
Q: What covers over 90\% of the Earth's surface and 0\% of the moon's surface\\
Answer Choices:\\
(a) a magnesium iron silicate mineral (b) chemical element with the symbol S (c) the element with the symbol Fe (d) that which contains 2 hydrogen and 1 oxygen molecules\\
\vspace{-1mm}
A: Water covers over 90\% of the Earth's surface and 0\% of the moon's surface, and contains 2 hydrogen and 1 oxygen molecules. So the answer is (d) that which contains 2 hydrogen and 1 oxygen molecules.\\
    \bottomrule
    \end{tabular}
    \caption{    
    Chain of Thought prompt for OpenbookQA.
    }
\end{table}

\begin{table}[H]

    \centering
    \small
    \begin{tabular}{p{7.8cm}}
        \toprule
        \vspace{-2mm}
Q: What is the most likely to be an effect of acid rain on an aquatic environment?\\
Answer Choices:\\
(a) decrease in plant life (b) increase in fish population (c) increase in plant growth (d) cleaner and clearer water\\
\vspace{-1mm}
A: (a) decrease in plant life \\
Explanation: Acid rain has a negative impact on water quality and decrease in water quality leads to a decrease in aquatic life.\\
~\\
Q: The moon's surface\\
Answer Choices:\\
(a) is smooth on the entire surface
(b) contains large cavities cause by explosions
(c) contains an internal core of cheese
(d) is filled with lakes\\
\vspace{-1mm}
A: (b) contains large cavities cause by explosions\\
Explanation: The moon's surface contains many craters and craters are large cavities caused by explosions.\\
~\\
Q: As a car approaches you in the night\\
Answer Choices:\\
(a) the headlights become more intense
(b) the headlights recede into the dark
(c) the headlights remain at a constant
(d) the headlights turn off\\
\vspace{-1mm}
A: (a) the headlights become more intense\\
Explanation: Headlights of a car are source of light. As a source of light becomes closer , that source will appear brighter.\\
~\\
Q: When the weather changes as it does from Christmas to Easter,\\
Answer Choices:\\
(a) the air may chill
(b) the ground may freeze
(c) the plants may die
(d) the ground may warm\\
\vspace{-1mm}
A: (d) the ground may warm\\
Explanation: Christmas is in the winter season while Easter is in the beginning of spring. Weather change from winter to spring is to become warm.\\
~\\
Q: Heat and moisture in the ocean is a good recipe for\\
Answer Choices:\\
(a) a violent storm
(b) violent sea animals
(c) condensation
(d) inland storms
\vspace{-1mm}
A: (a) a violent storm\\
Explanation: An ocean is a source of heat and moisture for a hurricane and a hurricane is a violent storm.\\
~\\
Q: Hummingbirds take what with them\\
Answer Choices:\\
(a) Bees
(b) energy
(c) Pollen
(d) Honey
\vspace{-1mm}
A: (c) Pollen\\
Explanation: When hummingbirds dip their long bills into flowers to drink nectar, they also pick up grains of pollen. Pollen that sticks to a hummingbird’s feathers and bill gets carried to the next flower it visits.\\
~\\
Q: What covers over 90\% of the Earth's surface and 0\% of the moon's surface\\
Answer Choices:\\
(a) a magnesium iron silicate mineral
(b) chemical element with the symbol S
(c) the element with the symbol Fe
(d) that which contains 2 hydrogen and 1 oxygen molecules\\
\vspace{-1mm}
A: (d) that which contains 2 hydrogen and 1 oxygen molecules\\
Explanation: Water covers over 90\% of the Earth's surface and 0\% of the moon's surface, and contains 2 hydrogen and 1 oxygen molecules.\\
    \bottomrule
    \end{tabular}
        \caption{    
    Rationalization prompt for OpenbookQA.
    }
\end{table}
\begin{figure*}[!t]
  \centering
  \includegraphics[scale=0.50]{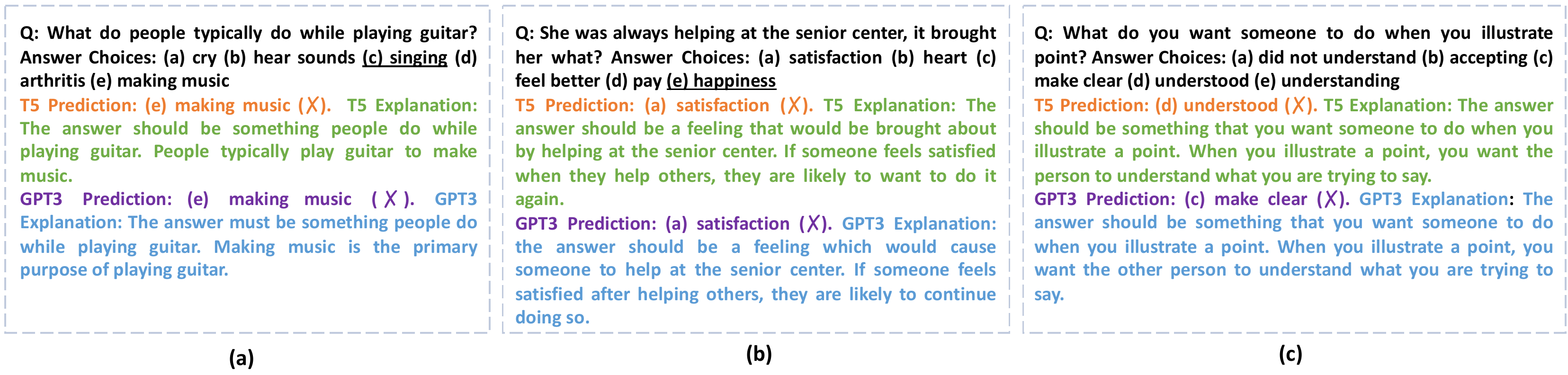}
  \caption{T5 and GPT-3 generated explanations with incorrect predictions.}
      \label{fig:incorrect_example}
\end{figure*}
\section{Explanation examples}  \label{appendix:examples}
Here we further provide three examples as shown in Figure \ref{fig:incorrect_example}, where both T5 and GPT-3 have incorrect predictions. We observe that in both example (b) and (c), T5 and GPT-3 have plausible predictions and explanations although their predictions are different from golden labels. 
\end{document}